# Decomposition of forging die for high speed machining

Laurent Tapie [1, 2], Bernardin Kwamivi Mawussi [1, 2]

**(1) :** Laboratoire Universitaire de Recherche en Production Automatisée ENS Cachan ; 61, avenue du Président Wilson ; 94235 Cachan cedex, France
*+(33)1 47 40 27 52/ +(33)1 47 40 22 20*
*E-mail :* {laurent.tapie, kwamivi.mawussi}@lurpa.ens-cachan.fr

**(2) :** IUT de Saint Denis, Université Paris 13; Place du 8 mai 1945 ; 93206 Saint Denis Cedex, France
*+(33)1 49 40 61 66/ +(33)1 49 40 61 76*

**Abstract:** Today's forging die manufacturing process must be adapted to several evolutions in machining process generation: CAD/CAM models, CAM software solutions and High Speed Machining (HSM). In this context, the adequacy between die shape and HSM process is in the core of machining preparation and process planning approaches. This paper deals with an original approach of machining preparation integrating this adequacy in the main tasks carried out. In this approach, the design of the machining process is based on two levels of decomposition of the geometrical model of a given die with respect to HSM cutting conditions (cutting speed and feed rate) and technological constrains (tool selection, features accessibility). This decomposition assists machining assistant to generate an HSM process. The result of this decomposition is the identification of machining features.

## 1- Introduction

Today, design of forged part and forging die is done in the engineering and design department. Die analysis is performed on the CAD model of die cavities with metal flow simulation [T1] [KM1]. Thus, die designers define forged part and dies according to CAD models and forging process simulation. The die model obtained is transferred to the machining process department. Recent die models have more complex shape. The introduction of close die forging process in order to limit re-machining operation (Fig. 1) increases geometrical complexity of die shapes.

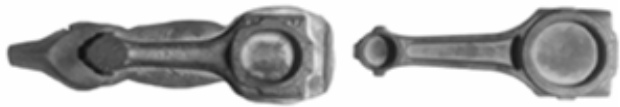

**Figure 1: Conventional versus close die forging process [DB1]**

Because of the evolution of die shapes, the tasks carried out by the machining assistant are different insofar as he must work out the machining process starting from die model given by the engineering and design department. This model is considered in most of the cases as a void geometry, since it involves poor technological and topological information required by machining process assistant. So, machining process preparation must be based on geometrical and topological die shape analysis to associate machining information. This information must be useful to choose machining strategy and resources according to die shape with respect to the minimal machining time, the part quality and surface roughness. Integration of High Speed Machining (HSM) technology is also crucial for the processing of technical and economical constrains.

Current forging die machining process is composed of three main steps: roughing, semi-finishing and finishing. Roughing operations objective is to reach as close as possible the desired finished shape with high level of cutting material flow and tool path series as smooth as possible. Semi-finishing operations objective is to warrant a constant depth of cut or constant tool engagement during finishing operation. Finishing operations objectives are to obtain the final die shape up to the geometrical quality specified. The die geometry and topology must be taken into account during these machining operations with a high leveled refinement to reach these objectives.

The approach proposed in this paper is limited to finishing operation of die cavity. In the industrial context, machining resources often used are 3-axes high speed machine tools associated with ball nose or corner rounded end mill. CAM strategies used during this operation are very common: parallel planes guidance, Z level guidance, parallel curve guidance.

Technical and economical studies highlighted the importance of CAM preparation in the die machining process [AL 1]. During this preparation step, the machining process assistant know how have a great influence on the part quality. Thus, the tasks carried out by the machining process assistant must be adapted to the technical and economical context which involves evolutions on CAD models, CAM strategies and machining process (HSM). This adaptation depends on the good adequacy between die shape and HSM process. Indeed, a bad adequacy involves increases in machining time and decreases in part quality. A well known cause of part quality and surface integrity decreases is the difference between real cutting conditions and HSM cutting conditions set point [LL1] [WK1] [MH1]: keep up a constant feed rate (deceleration and high breakages), avoid high tool engagement variation, keep up a constant cutting speed (avoid small effective cutting diameter).

To solve this problem the work presented in this paper proposes to decompose the die geometrical model in order to adapt locally a HSM process to die shape. This decomposition is based on the difficulties involved by HSM technology.

## 2- Proposed approach

Tasks carried out by the machining assistant are generally based on CAM/CAPP software modules and machining feature concept. Nevertheless, few works are focused on complex shape machining features. The approach presented in this paper is in keeping with the general pattern of these features creation [BB1].

The main objective of the work presented in this paper is to define machining features suitable for HSM process starting from the die model (3D CAD model). The CAD model decomposition based on difficulties induced by HSM process is an interesting concept to obtain a good adequacy with the part geometry. This concept is already used in the project "Usiquick" [U1]. The proposed approach is carried out starting from the analysis of the difficulties related to the machining of die shapes (Fig. 2). Then, results obtained at this level are superposed to obtain machining features and their topological relationship. A machining process is generated for each machining feature and then all the local machining processes are adapted for the entire forging die. This paper is focused on the topological decomposition of die model from machining difficulties.

## 3- Decomposition method

The CAD model decomposition for machining features creation is very efficient. Besides, machining difficulties criteria are very relevant as it is highlighted in [DC1].

The die shape is decomposed in machining difficulties features. These features helped the machining process assistant to identify machining features. Two types of HSM process difficulties encountered are studied: to keep the set point of the cutting speed and to find better sequences of machined surface suitable for the application of standard machining strategies.

### 3.1 – Cutting speed respect

As it is highlighted in several academic papers, to maintain the cutting speed set point is a key factor in the cutting mechanism. Indeed, thermal variations and cutting force variations in the cutting feature are observed when the effective cutting speed is significantly reduced. These variations can induce premature tool wear [T2], dimensional error, bad surface roughness [T2] and part metallographic variation [WK1].

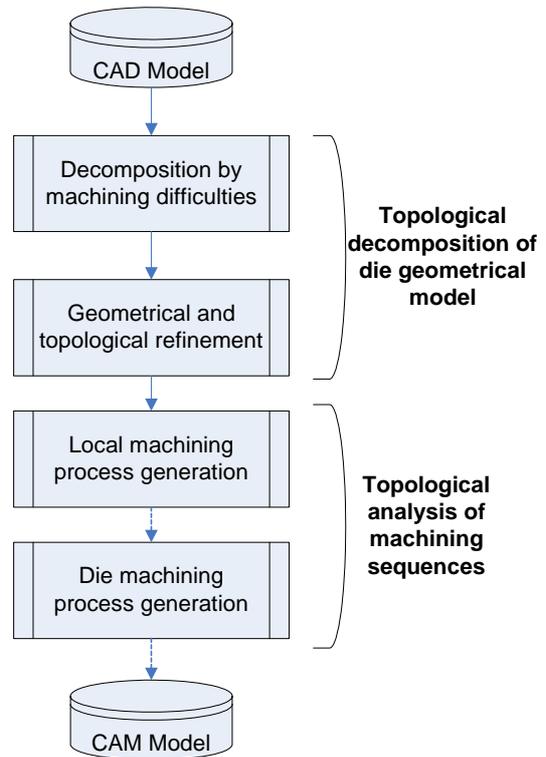

**Figure 2: Structure of the decomposition process**

The effective cutting speed is proportional to the effective diameter of the cutting tool. During finishing operations of complex shapes or sculptured surfaces with ball-end mill and corner rounded end mill with a large corner radius, the effective diameter varies permanently. To avoid calculating the effective cutting speed at any point, its variation can be associated to different contact areas defined by Lamikiz & al. (Fig. 3) [LL1 & 2]. The critical feature for a ball end mill is the centre. In this feature the cutting speed is quasi null and the chip evacuation is critical.

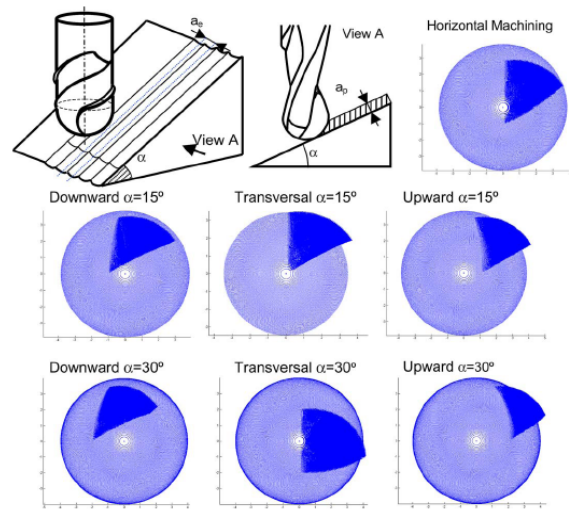

**Figure 3: Variations of the contact area between tool and part [LL1 & 2]**

### 3.1.1 – HSM of forging die analysis

HSM of forging die is very often performed in a 3 axis machine tool associated with a ball end mill. Two machining strategies widely developed in CAM software are often used: the iso-plane tool path generated by the slicing approach parallel (parallel planes strategy) or perpendicular to the tool axis (Z-level strategy). These strategies are analysed according to the contact area between tool and part.

The parallel planes analysis shown in Fig. 4 highlights the variations of the tool and part contact area along a given tool trajectory. For the tool trajectory illustrated Fig. 4.a, the contact area is always close to the tool tip. This specific cutting position induces a difficult chip evacuation, uncut material (metal pick-up) and premature tool wear. Yet, the tool and part contact area shape is quasi-constant. For tool trajectory illustrated Fig. 4.b, the tool and part contact area shape varies along the tool path. Thus, tool alternate machining configuration close to the tip and close to the flank with up and down milling. Several phenomena occur in this case: tool bending, tool extraction, mechanical alteration of the part shape [BM1] [MH2].

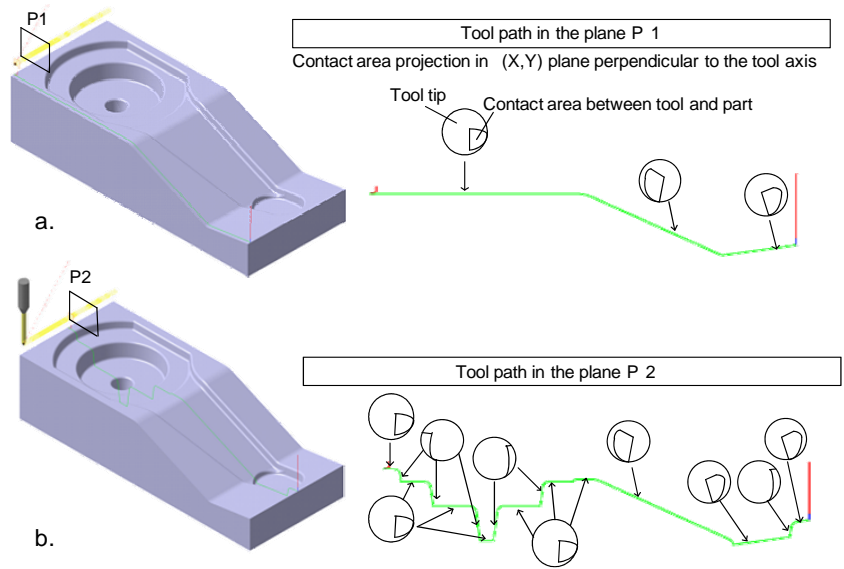

**Figure 4: Tool and part contact area analysis for parallel planes strategy**

For the toolpath generated according to the Z-level strategy, Fig. 5, the tool and part contact area shape and position are constant in the whole part. Yet, in some features the tool engagement is higher, particularly when the part shape curvature is equal or lower than the tool radius.

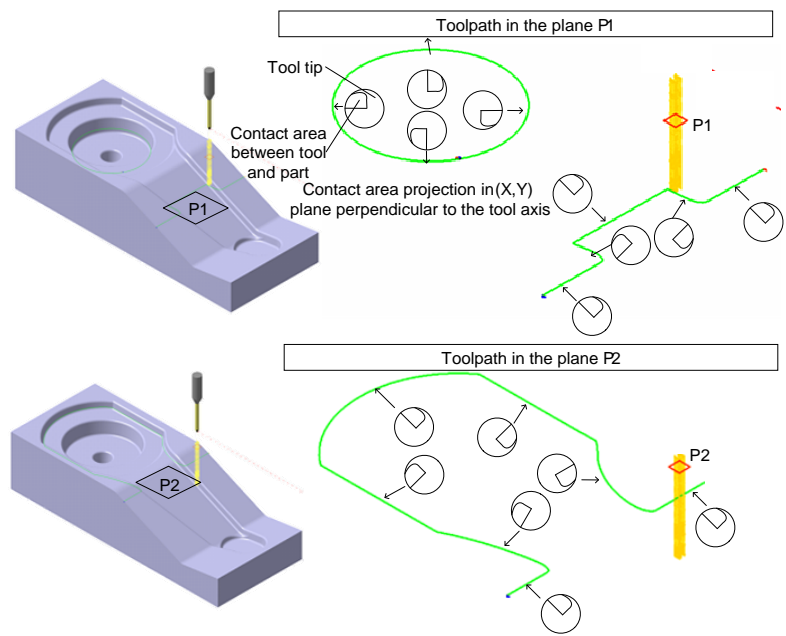

**Figure 5: Tool and part contact area analysis for Z-level planes strategy**

These two analysis highlight the more significant causes of machining process or part shape alterations: no respect of cutting speed and some phenomena linked with the toolpath choice.

The two iso-plane tool path generated by the slicing approach for the whole part highlight the difficulty to limit the effective cutting speed variations.

### 3.1.2 – Contact area type

The analysis of 3-axes machining of forging dies makes it possible to define two types of contact areas: end contact and flank contact (Fig. 6). The first type of contact area "end contact" is located to the tool tip. A great reduction of effective cutting speed occurs in this feature because of the proximity of the spindle axis. As the second type of contact area "flank contact" is located in the periphery of the cutting tool. A cutting speed close to theoretical one can be reached.

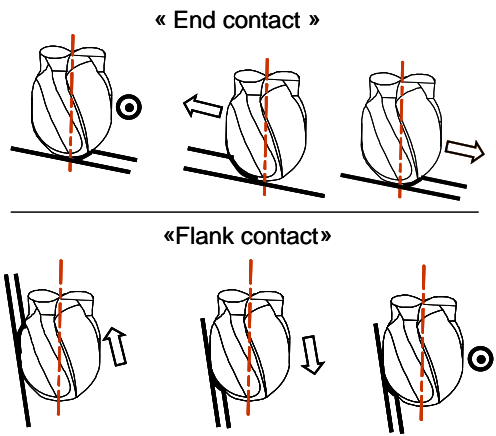

**Figure 6: Contact area type**

The contact area type is linked to the inclination angle β between tool axis and part material direction (Fig.7). Thus, this contact area type is analysed using this angular parameter.

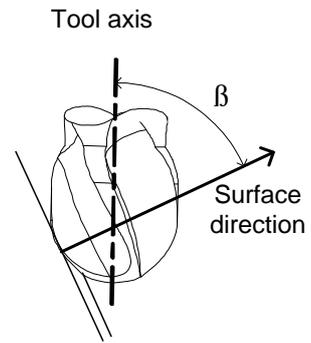

**Figure 7: Contact area type analysis**

### 3.2 – Cutting speed variation

CAD forging die model are very often designed with a B-Rep model. Nevertheless, the data used in this kind of models (vertex, edge, surface) are complex to be manipulated by machining assistant. This complexity is highlighted in the Usiquick Project [D1]. Geometrical and topological refinement of a B-Rep model involves designing complex algorithm in order to recognize automatically pre-defined machining features.

To simplify the B-Rep data in our approach, the forging die B-Rep model is converted in a polygonal model. The STL model is chosen because it is a very common model associated with computer aided conversion algorithm in several CAD, CAM and CAPP software.

Starting from the STL model we generate a map which represents the cutting speed variation. The main objective of this map is to search shapes having the same type of contact areas between the cutting tool and the part. Then the map is obtained by colouration of each facet according to the criterion δ (Fig. 8, δ=cos(β), β angle between tool axis direction and material direction). The tool axis is definied according to 3 axis machining. Thus the cutting tool axis is colinear to the Z axis of the process. The material direction defines the free material side of the shape. Each material direction is given by the corresponding STL facet normal.

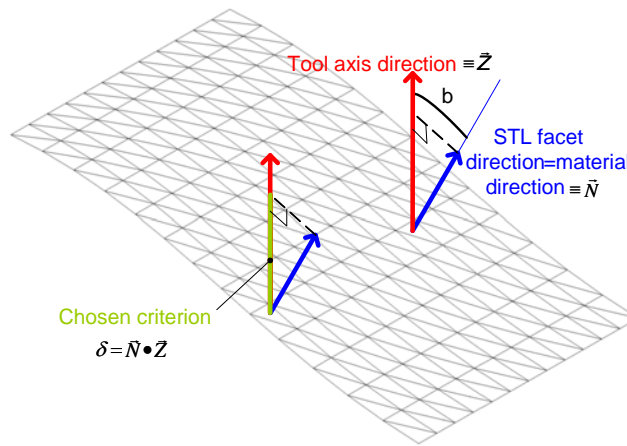

**Figure 8: Cutting speed map construction**

According to the analysis of several industrial dies different machining features can be extracted. The example shown in Fig. 9 illustrates these features :

- In dark red unique shade, $\delta = 1$, the tool tip is used ("end contact" configuration), the contact area between the tool and the part is constant. This type of contact induces a low effective cutting speed associated to a ball end mill tool. This feature type is qualified as "horizontal" feature.

- In red, $\delta \approx 1$, the tool tip is used ("end contact" configuration Fig. 6). This feature type is qualified as "quasi-horizontal" feature. The limiting threshold with a "flank configuration" is fixed by the machining assistant. Yet, the colour map allows to visualize the shapes which have the same contact area. In Fig. 9, two "horizontal" features are identified at the die top ($\delta = 1$, dark red) in relation with a red feature, $\delta \approx 1$. Thus, this feature is necessarily "quasi-horizontal".

- In deep blue unique shade, $\delta = \cos(f)$, $f$ is the die draft angle, the tool flank is used ("flank contact" configuration), the contact area between the tool and the part is constant. This type of contact induces an effective cutting speed close to the required cutting speed. This feature type is qualified "draft" feature.

- In blue, $\delta \approx \cos(f)$, the tool flank is ued ("flank contact" configuration). The limiting threshold with a "flank configuration" is decided by the machining assistant. This feature type is qualified "quasi vertical" feature.

- In colour gradation between red and blue, the tool is sollicited with the "end contact" and "flank contact" configuration. These features appear between "horizontal" or "quasi-horizontal" feature and "draft " or "quasi-vertical" feature. In this case, colour graduation features are qualified "transition" features.

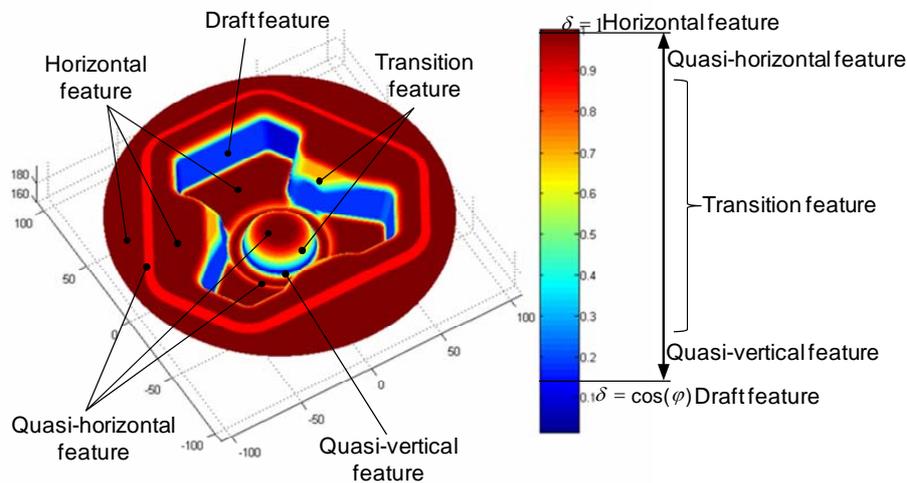

**Figure 9: Cutting speed distribution map**

### 3.3 – Sequences of machined surfaces

Several works are related to inclined plane machining [KK2] [KK3] [T3] [T4] or complex shape part machining [BA1] [VL1]. They underline the fact that surfaces must be grouped according to the same direction series in order to limit cutting tool material engagements and exits. Indeed, each cutting tool engagement or exit in the part induces mechanical and thermal problems.

Three types of sequences of machined surfaces can be defined in the case of 3 axis forging die machining (Fig. 10). They are defined according to the tool axis direction and the material direction.

For the "simple sequence" type, a privileged planar direction P (Fig. 10.a) perpendicular to the tool axis direction allows connecting surfaces together which have the same material direction and are topological adjacent.

For the "oriented sequence" type, a privileged planar direction P (Fig. 10.b.) containing the tool axis direction allows connecting surfaces together which have the material direction contained in this specified plane and are topological adjacent.

For the "indifferent series" type any privileged planar direction can be found (Fig. 10.c.).

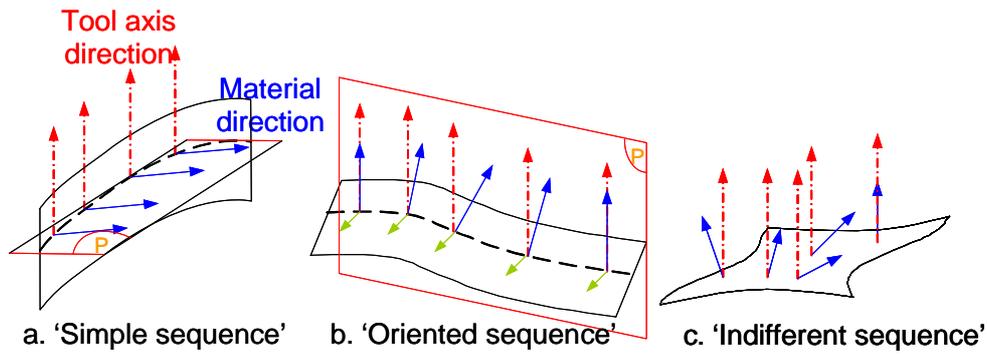

**Figure 10: Surfaces sequence types**

### 3.4 – Surface sequences associated maps

The main objective of these maps is to identify part features corresponding to the same surface sequence type. According to these maps, privileged directions for the parallel planes strategy are given to machining process assistant. These maps are obtained by colouration of each facet of the STL model according to the three sequence types previously definied.

The first map is designed for the identification of surface sequence in planes perpendicular to the tool axis which are associated to the Z-level strategy (Fig. 11). This map is already obtained by the same way from the cutting speed map. Nevertheless, the map interpretation is different. "Simple" sequences are obtained only if the criterion δ (δ=cos(β) is constant for the whole connected surfaces.

To identify the "oriented sequences", several directions are evaluated (Fig. 12). For a given direction, each facet is colourised according to the criterion of the angle χ in Fig. 12 between the fictive direction (line in green Fig. 12) and the material direction projection (doted line in blue Fig. 12) in the plane normal to the tool axis (plane P, Fig. 12). The material direction is given by the normal of the STL facet (Fig. 12).

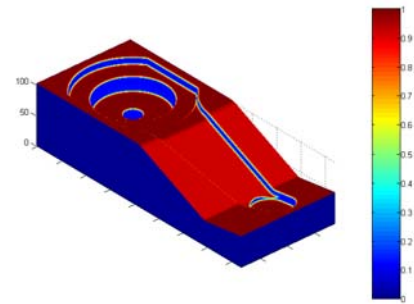

**Figure 11: Simple sequence associated map**

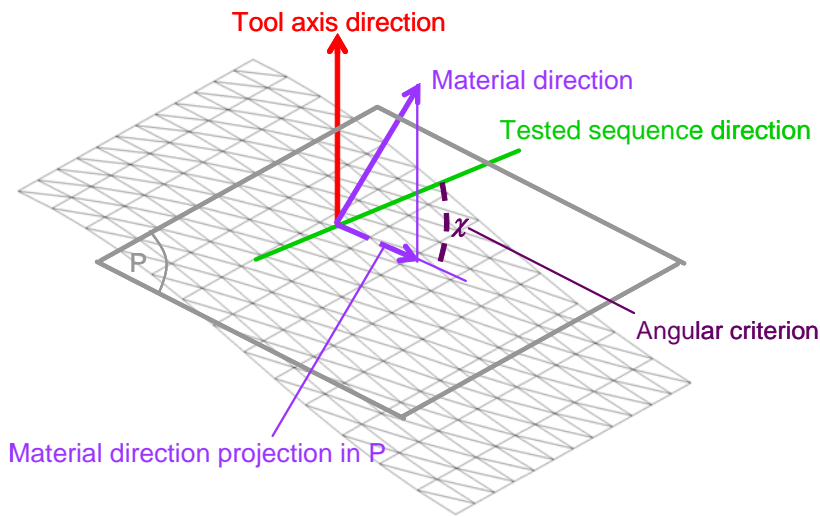

**Figure 12: Oriented sequences associated map construction**

In the examples (Fig. 13), ten tested directions are mapped. The direction range is 90° with a step of 10°. This choice is done by machining assistant. The step can be refined around pertinent direction.

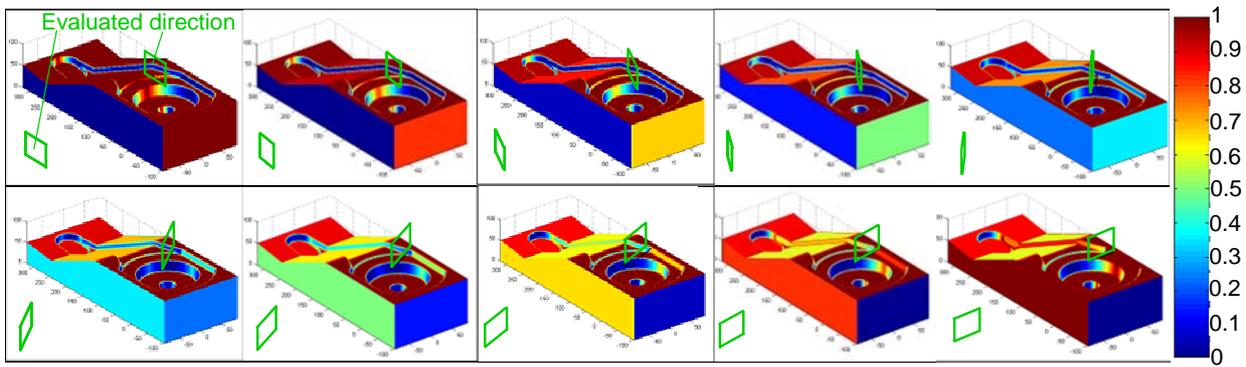

**Figure 13: Oriented sequences associated maps examples**

## 4- Decomposition maps exploitation

Machining process assistant has to choose machining equipments (machine tools and cutting tools) and CAM machining strategies to compute tool trajectories during machining process preparation. Thus, he needs some information to make these choices. These infromation are given by the decomposistion maps. This paragraph highlights the decomposition maps exploitation and complementarity's in order to help machining process assistant to make choices.

### 4.1 – Cutting tool choice

The cutting speed distribution map is used to associate identified machining features with cutting tool (Fig. 14).

For instance, ball end mills tools are associated with "drafted" and "quasi-vertical" features. During the machining of these features, the effective cutting speed is closed to theorical one.

Ball end mills are also associated with "transition" features to limitate tool trajectories compared with corner end mill.

For "horizontal" features end mill or corner end mill are associated. These tool types allow to keep an effective cutting speed close to the theorical one. Yet, in the case of corner end mill use the torus radius must be small to limitate cutting speed decrease.

For "quasi-horizontal" features, corner end mill are associated for the same reason for "horizontal" features. Besides, if a "quasi-horizontal" feature is composed of a portion of "horizontal feature", ball end mill must be excluded because of a too small effective cutting radius.

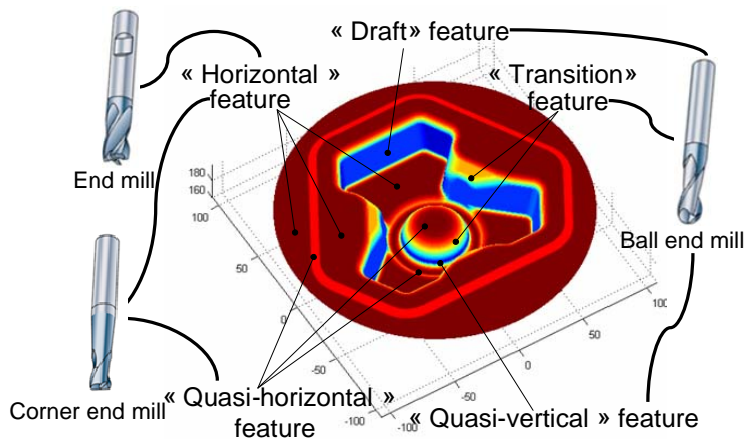

**Figure 14: Cutting tool association**

### 4.2 – Machining strategy choice

The machining strategy choice associated with decomposition maps (cutting speed distribution, oriented sequences) is focus on planar machining strategies as it is presented previously. Indeed, these machining strategies are the core of the majority of CAM software. Besides, these strategies are often the first choices done by machining assistant because of they are easy to generate.

For "drafted" and "quasi-vertical" features Z-level strategy is associated (Fig. 15). These features are machined by Z-level strategy limiting strong variations of tool and part contact area. For the same reasons "transition" features are also associated with Z-level strategy (Fig. 15).

For "horizontal" features a surfacing machining strategy is associated (Fig. 15).

For "quasi-horizontal" features information given by cutting speed distribution associated map may be crossed with information given by oriented sequences associated maps. If a "quasi-horizontal" feature is "indifferent" any privileged machining direction is needed thus a surfacing strategy is associated (Fig. 15). If a "quasi-horizontal" feature is "oriented" a privileged vertical planar machining direction is suitable thus a parallel plan strategy is associated in the privileged oriented vertical plan (Fig. 15).

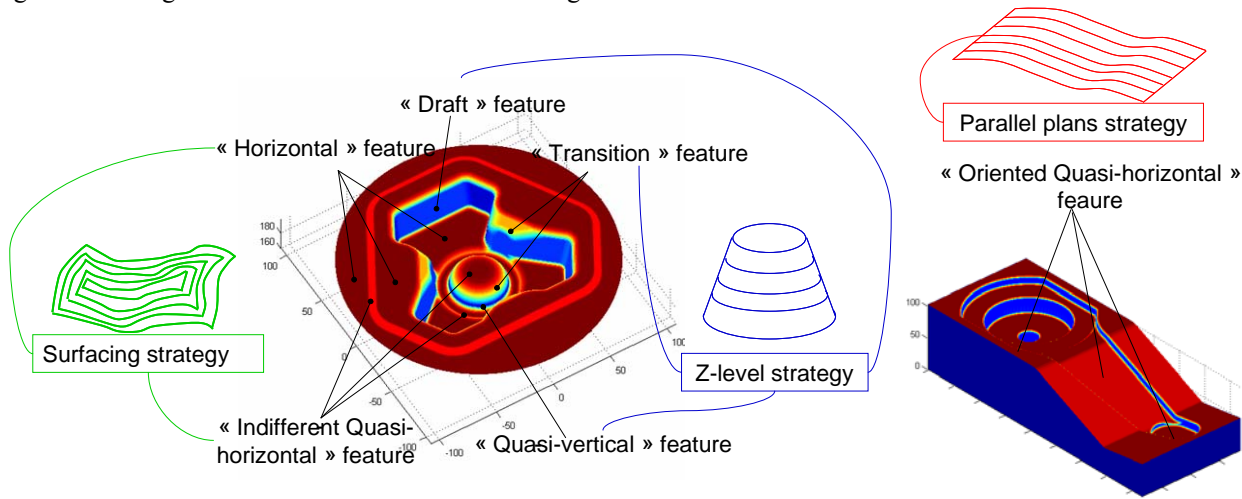

**Figure 15: Machining strategy association**

## 5- Conclusions

Today's forging die manufacturing approach must be adapted to economical choices which involve evolution and substantial modification in die machining preparation, die design and CAD models, CAM software, machining process HSM. Thus, the adequacy between part geometry and HSM process must be the core of machining process assistant approach.

In this paper, we proposed a new approach intented to help machining process assistant during HSM process generation. This approach is based on the decomposition of forging die geometrical model and topological analysis of machining sequences. The topological decomposition is based on machining process difficulties decomposition concept. This approach is particularly focused on HSM cutting conditions respect (cutting speed, feed rate). Otherwise, the geometrical model refinement is another important concept used in our approach. Thus, according to these concepts the machining process assistant has machining information linked with the die geometrical model. He can use these information to generate the die cavity HSM process during the topological analysis of machining sequences.

The die topology being the core of HSM process difficulties, the CAD model is decomposed according to specific topological information extracted from machining difficulties presented by maps: cutting speed map and machining direction sequences maps. Then these maps are superposed to decompose geometrical die model and extract machining features.

According to several industrial forging die studied with this topological decomposition five machining features are defined: flank feature, simple floor feature, oriented floor feature, indifferent floor feature, and transition feature. Yet, the boundaries between machining features are not automatically identified. Machining assistant has to make the correspondence between the STL and B-Rep CAD model.

Topological decomposition by machining difficulties of freeform parts is appearing to be an efficient systematic approach for machining process assistant. This new concept will be presented in the future.

## 6- References